\documentclass[manuscript,screen,sigconf]{acmart}
\setcopyright{none}
\renewcommand\footnotetextcopyrightpermission[1]{}
\AtBeginDocument{%
  \providecommand\BibTeX{{%
    \normalfont B\kern-0.5em{\scshape i\kern-0.25em b}\kern-0.8em\TeX}}}

\newcommand{\thetitle}{Human Response to an AI-Based Decision Support System:  A User Study on the Effects of Accuracy and Bias}
\newcommand{\theauthorspdfinfo}{Anonymous}

\usepackage{amsmath}
\usepackage{amsfonts}
\usepackage{graphicx}
\usepackage{natbib}
\usepackage{multirow}  % multi-row cells in tables
\usepackage{xcolor}    % color definitions
\usepackage{paralist}  % \inparaenum command and others
\usepackage{xspace}    % \xspace command
\usepackage{booktabs}  % better tables
\usepackage{makecell}
\usepackage[inline]{enumitem}

\newcommand{\spara}[1]{\smallskip\noindent\textbf{#1}}
\usepackage[hide]{chato-notes}
\title{\thetitle}

\begin{document}


\author{David Solans}
\email{david.solans@upf.edu}
\orcid{0000-0001-6979-9330}
\affiliation{%
  \institution{University Pompeu Fabra, Barcelona \country{Spain}}
}

\author{Andrea Beretta}
\email{andrea.beretta@isti.cnr.it}
\affiliation{%
 \institution{National Research Council, Pisa \country{Italy}}
}

\author{Manuel Portela}
\email{manuel.portela@upf.edu}
\affiliation{%
  \institution{University Pompeu Fabra, Barcelona \country{Spain}}
}

\author{Carlos Castillo}
\email{chato@acm.org}
\affiliation{%
  \institution{University Pompeu Fabra, Barcelona \country{Spain}}
}

\author{Anna Monreale}
\email{annam@di.unipi.it}
\affiliation{%
  \institution{University of Pisa, Pisa \country{Italy}}
}

%\author{ANONYMOUS AUTHORS}
%\renewcommand{\shortauthors}{D.Solans, et al.}

\begin{abstract}
Artificial Intelligence (AI) is increasingly used to build Decision Support Systems (DSS) across many domains.
This paper describes a series of experiments designed to observe human response to different characteristics of a DSS such as accuracy and bias, particularly the extent to which participants rely on the DSS, and the performance they achieve.

In our experiments, participants play a simple online game inspired by so-called ``wildcat'' (i.e., exploratory) drilling for oil.
%
%The game is played in a 32x32 grid and the goal is to maximize the total score after various rounds. %, balancing exploration and exploitation on a complex landscape.
%
The landscape has two layers: a visible layer describing the costs (terrain), and a hidden layer describing the reward (oil yield).
The final score of a participant is computed as rewards minus costs.
%The terrain determines drilling costs and the oil yield determines income, with the final score being income minus costs.
%
Participants in the control group play the game without receiving any assistance, while in treatment groups they are assisted by a DSS suggesting places to drill.
For certain treatments, the DSS does not consider costs, but only rewards, which introduces a bias that is observable by users.
Between subjects, we vary the accuracy and bias of the DSS, and observe the participants' total score, time to completion, the extent to which they follow or ignore suggestions. We also measure the acceptability of the DSS in an exit survey.
%
%To conduct the study, which was approved by our Ethics Review Board, we used a crowdsourcing platform to recruit more than 400 participants. 

Our results show that participants tend to score better with the DSS,
that the score increase is due to users following the DSS advice,
and related to the difficulty of the game and the accuracy of the DSS.
%
%Additionally, we observe  that they take more time to play when there is a DSS.
%
We observe that this setting elicits mostly rational behavior from participants, who place a moderate amount of trust in the DSS and show neither algorithmic aversion (under-reliance) nor automation bias (over-reliance).
However, their stated willingness to accept the DSS in the exit survey seems less sensitive to the accuracy of the DSS than their behavior, suggesting that users are only partially aware of the (lack of) accuracy of the DSS.
The game constitutes a research platform intentionally designed to study decision support in the absence of pre-existing expertise, which makes it an interesting model for studying algorithmic reliance.
%
%The platform is available as free software for future research.

\end{abstract}




\maketitle
\pagestyle{plain}
\thispagestyle{empty}

% This figure should be at the top of the second column
% of the first page, this is customary in HCI papers.
%
% It should also be referenced from the first page,
% although that might be tricky.
% -- ChaTo
\begin{figure}[ht] 
    % Not sure about how much to trim -- ChaTo
    %\centering\includegraphics[trim={0 0 80px 175px},clip,width=0.9\columnwidth]{Figures/game_UI.png}
    \centering\includegraphics[trim={5px 0 20px 5px},clip,width=0.9\columnwidth]{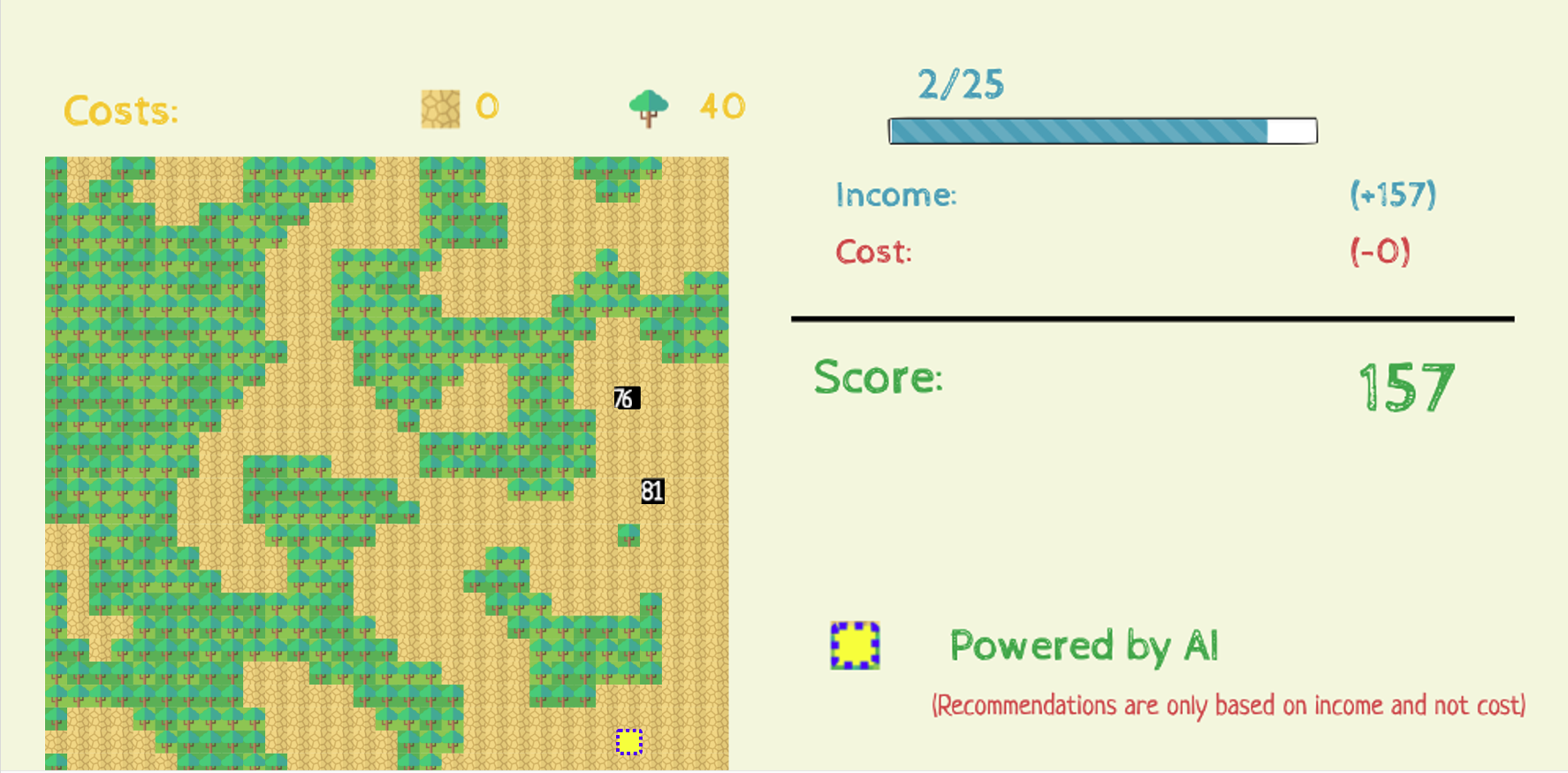}
    \caption{Experiment interface. The terrain is represented by green (forest) and brown (desert) cells.
    \label{fig:experiment-interface}
    The user has drilled in the cells in black, and a recommendation from the DSS (in yellow) is shown.}
\end{figure}

\section{Introduction}
\label{sec:intro}

Decision support systems based on artificial intelligence are increasingly being deployed in a variety of real-world scenarios \cite{lotlikar_2021,Bogen2018HelpWA, Mukerjee_2002,Angwin_2016}. 
% 3-5 citations. Medicine, Hiring, Bank loans, Criminal Justice
%
Typically, these deployments involve a moderate level of automation~\cite{Cummings04automationbias} in which a human is in charge of making the final decision based on some inputs, including a Decision Support System (DSS).
Indeed, in some contexts, that a human makes the final decision when using a decision support system might increasingly become a legal mandate.\footnote{See, e.g., Article 22 of the EU General Data Protection Regulation.} %for a broad range of highly consequential decisions human decision-making or at least oversight might increasingly be a legal mandate.

The process of incorporating the input from a DSS is a complex cognitive task, considering that in many domains machines are still far from achieving perfect accuracy.
Users have to navigate between the traps of algorithmic aversion~\cite{Dietvorst_2014}
and automation bias~\cite{Cummings04automationbias},
respectively characterized by under- and over-reliance on the DSS.
Several authors have studied and tested methods to increase user trust in machine predictions \cite{Balfe2018UnderstandingIK};
however, %trust is an elusive context, and more importantly,
eliciting more trust might not be the best way to combine human and machine intelligence.
Arguably, the ideal scenario is one in which users follow the advice of the DSS when it is correct and ignore it when it is wrong.

We study how key characteristics of a decision support system impact the decisions made.
To do this, we designed an online game, depicted in Figure~\ref{fig:experiment-interface}.
It consists of a 32x32 board representing a map having green (``forest'') and brown (``desert'') cells. 
Players have to ``drill'' for oil by clicking on a cell, and obtain a score equal to the oil yield of that cell, which is revealed only after clicking, minus the ``environmental cost,'' which is zero for desert cells, but non-zero for forest cells.
The hidden oil profile is independent of the terrain map, and is such that neighboring cells offer a similar reward.
The goal of the participants is to maximize the score after a series of 25 rounds (clicks) in each of three maps having different levels of difficulty.
Participants in our control group do not receive any assistance, while those in various treatment groups are assisted by decision support systems of varying levels of accuracy and bias.

\spara{Research question.}
Our main research question is \textit{how do the characteristics of a decision support system impact human performance, time to completion, and reliance?}
We address this research question experimentally.

\spara{Our contribution.}
We release the design and implementation of a platform to study the stated research question.
As a limitation (discussed on \S\ref{sec:conclusions}), the platform does not simulate a high-stakes scenario.
However, it allows studying common elements of human interaction with a DSS, as it does not require prior knowledge from participants.
In this platform, researchers can vary the problem difficulty and the accuracy of the decision support system, and introduce bias in the recommendations in a manner that is visible to participants.
The platform records all interactions, generating a wealth of data including player's performance, timing, and reliance on the decision support system both implicitly (by observing clicks) and explicitly (by an exit survey).

We describe a series of experiments, approved by our Ethics Review Board, and involving over 400 participants recruited via crowdsourcing.
These experiments uncover what we consider mostly rational behavior.
For instance, participants rely on the decision support system to an extent that is well-aligned with its accuracy.

This paper is organized as follows:
We overview related work (\S\ref{sec:related-work}) before describing the methodology (\S\ref{sec:methodology}) and experimental design (\S\ref{sec:experimental-design}).
Then, we review the obtained results (\S\ref{sec:results}) and discuss the general findings (\S\ref{sec:discussion}).
Finally, we discuss the limitations of our approach and future work (\S\ref{sec:conclusions}).

\section{Related work}
\label{sec:related-work}

The literature on decision support is vast; in this section, we overview research on Decision Support Systems (DSS) that provides context for our work (\S\ref{subsec:relwork-dss}), particularly research on trust and reliance on DSS (\S\ref{subsec:relwork-trust}), and on the influence of DSS accuracy on reliance (\S\ref{subsec:relwork-accuracy}).

\subsection{Decision Support Systems (DSS)}
\label{subsec:relwork-dss}

Decision-making is an essential activity that involves facing choices, often in the presence of uncertainty~\cite{Bucinca2020}.
It is also a complex cognitive process that depends on interpreting large amounts of information, evaluating the possible consequences of the decision to be made~\cite{Cummings04automationbias}. 
The goals of deploying a DSS to assist human decision-making typically include improving the quality or accuracy of decisions, reducing subjectivity, reducing costs, and increasing the efficiency of a decision-making process~\cite{Grgic-Hlaca2019, larus2018computers}.

An important characteristic of a DSS is its degree of autonomy with respect to human decision makers~\cite{Cummings04automationbias}.
A high level of automation, e.g., a DSS that can automatically implement its recommendations, can be useful in scenarios where the workload is high and the DSS can correctly make decisions in a reliable manner, helping to reduce workload~\cite{Balfe2018UnderstandingIK}.
A lower level of automation, e.g., a DSS that only recommends a choice but does not act upon it, might be less useful in some scenarios, but can also help human decision-makers detect failures or errors in the DSS~\cite{Tatasciore2019}.

One of the most challenging aspects of implementing a DSS is that humans embed values, biases, and assumptions in their decision-making without acknowledging the ambiguity, incompleteness, and uncertainty that are part of this process.
Hence, there can be misalignment or complete lack of alignment between the recommendations of a DSS and what humans would choose~\cite{Birhane2021, Tolan2018}.

\subsection{Trust and Reliance on Decision Support}
\label{subsec:relwork-trust}

Multiple studies from different disciplines have been carried out to understand which factors affect human trust and reliance on a DSS. 
Some controlled user studies indicate that participants tend to exhibit \emph{automation bias}, i.e., a tendency to follow the DSS even in cases in which they could have made better decisions by ignoring it~\cite{Dressel2018, Bansak2019, Suresh2020}.
Other controlled user studies have uncovered \emph{algorithmic aversion,} i.e., a tendency to distrust the recommendations of a DSS, or a rapid drop in confidence on an algorithmically-supported DSS after seeing it make a mistake, which would not have been of the same magnitude if the decision support were offered by a human~\cite{Dietvorst_2014}.

In general, the trust that users place in an automated system is affected by contextual, cultural, and societal factors~\cite{Lee2004}.
\emph{Trust} in this case is a complex construct that depends on the interplay of the users' disposition to the system, the situation in which the interaction happens, and what the users can learn about the system.
\emph{Reliance}, on the other hand, is more narrowly defined as compliance with an automation's recommendation~\cite{Hoff2015}. 
The extent to which trust determines reliance on a system is also subject to influences such as the complexity of the situation, its novelty, the degree of decisional freedom of the user, and whether s/he can compare the performance of automated and manual decisions~\cite{Hoff2015}.

In general, decision-makers consider the guidance of a DSS relative to the information context in which it is provided.
%
%Even with sufficient information about the system humans can
Hence, they may deviate from the DSS suggestions for different reasons, including their own biases, preferences, and deviating objectives~\cite{jahanbakhsh2020experimental,Green2020a,Mallari2020,stevenson2021algorithmic}.
For instance, some user studies suggest that experienced decision-makers in a given domain are less inclined to follow algorithmic suggestions, and rely more on their own cognitive processes \cite{Green2020,portela2022comparative}.

\subsection{Effects of DSS Accuracy}
\label{subsec:relwork-accuracy}

While a DSS does not have to be perfectly accurate for it to be useful~\cite{Acharya2018}, better decisions can be made if decision-makers can rely more on a more accurate DSS than on a less accurate DSS.
Some user studies indeed have shown that humans rely on machine predictions more when they are correct than when they are incorrect \cite{Lai2019a}.
However, in other user studies participants have consistently followed incorrect recommendations even for tasks they perform well~\cite{Suresh2020};
or have failed to correctly evaluate the accuracy of the DSS and their own accuracy, and hence have not been able to adapt their reliance on the DSS to its performance~\cite{Green2019,Green2019a}.
In general, perceiving the accuracy of a DSS is easier when DSS errors are consistent and deterministic, and when there is a simple boundary separating cases in which the DSS is correct from cases in which the DSS is incorrect~\cite{Bansal2019}.

Inferences about the accuracy of a DSS might be influenced not only by the correctness of the recommendations, in cases where humans can to some extent directly observe correctness, but also by the information provided by the DSS~\cite{Lai2019a}.
For instance, simply stating that a DSS is accurate can increase reliance on it, up to a point; however, the effect of these statements is weaker than direct observation of correct recommendations by users~\cite{yin_2019}.
Similarly, displaying a confidence score accompanying each recommendation or prediction of a DSS has been shown to increase willingness to rely on these recommendations, when the confidence score is high~\cite{Zhang2020}; however, in other user studies even when informing users that the DSS has low confidence in a recommendation, users have followed it~\cite{Suresh2020}.

We build upon previous work by studying a common interaction sequence with a DSS:
\begin{inparaenum}[(1)]
\item the environment provides an input,
\item a DSS recommends an action,
\item the human makes a decision, and
\item the environment returns a reward~\cite{Bansak2019}.
\end{inparaenum}
We provide a simple yet expressive scenario for which no prior experience from the participants is required.
With few exceptions (e.g.,~\cite{Suresh2020}) studies on the interaction of AI with human decision making do not experiment with task difficulty,
and in contrast with most previous work (e.g.,~\cite{Green2019a, Lin2020, Bansak2019, Dressel2018, Lai2019a, yin_2019}), the systems we consider are less accurate than humans acting alone.
To the best of our knowledge, ours is the first study that considers different levels of bias in the decision support.
\section{Methodology}
\label{sec:methodology}

In this section we describe the details of our methodology.
We first present an overview of our methodology (\S\ref{subsec:methodology-overview}) and the platform we designed to perform our experiments (\S\ref{subsec:platform}); then we describe the independent  %(\S\ref{subsec:independent-variables})
and dependent %(\S\ref{subsec:dependent-variables})
variables we take into consideration in our experiments (\S\ref{subsec:variables}).

\subsection{Overview}
\label{subsec:methodology-overview}

Our methodology is experimental and based on a simple yet expressive game, depicted in Figure~\ref{fig:experiment-interface}.
The game, inspired by ``Wildcat Wells''~\cite{Mason764} and described in~\S\ref{subsec:platform}, has several characteristics that make it appropriate for this research, including that
\begin{inparaenum}[(i)]
\item it is simple,
\item does not require training or prior experience,
\item uses a random game generator,
\item provides fine-grained control over game difficulty,
and
\item naturally lends itself to decision support.
\end{inparaenum}

We first perform a simple experiment without any decision support system, in which we randomly generate a set of maps and select three maps that (according to the scores users obtain) are labeled respectively as easy, medium, and hard.
Then, we experiment with these three maps by %and 
providing machine assistance in the form of a recommendation on where to click next, varying parameters such as the accuracy of the decision support or the amount of observable bias it might have.
We also include an exit survey in which we ask questions related to algorithmic reliance.

Participants are recruited through a crowdsourcing platform specialized in research\footnote{Prolific (\url{https://prolific.co/})} and paid above the platform-recommended fee of 7.5 GBP per hour of work.

\subsection{Platform}\label{subsec:platform}

This section describes the platform we designed and implemented to conduct experiments.\footnote{All of the code of the experimental platform will be released as free/open-source software with the camera-ready version of this paper.}

\spara{User interface.}
The interface is a web application composed of five screens:
\begin{inparaenum}[(i)]
    \item informed consent,
    \item demographic questions,
    \item tutorial,
    \item game, and 
    \item survey.
\end{inparaenum}
The informed consent form %approved by our Ethics Review Board,
explains the purpose, duration, risks and benefits of the experiment, and asks for explicit consent to participate.
Then, users are asked optional socio-demographic questions: \begin{inparaenum}[(i)]
    \item gender including male, female, and other;
    \item age bracket in five years increments;
    \item country of residence;
    \item level of education; and
    \item professional background.
\end{inparaenum}
Then, a short tutorial is shown to explain the gameplay.

The core portion of the experiment is the game, which consists of three maps shown in random ordering. Each of the three maps has a different level of difficulty: ``easy'', ``medium'', or ``hard''.
These maps were selected from a collection of candidate maps  generated using Perlin-noise~\cite{Perlin_2002} random generators for the terrain and oil profile.
The selection was performed through a preliminary experiment in which participants were asked to play the game without any assistance, as we explain in \S\ref{subsec:map-selection}. % for more details on the preliminary experiment for maps selection.

In each 32x32 map, the game proceeds in 25 rounds, i.e., allowing the user to click on 25 of the 1,024 cells. 
The total score of a user is income minus costs. 
The income is the combined ``oil'' yield of the 25 selected cells, which remains hidden until the cell is clicked.
The cost is a fixed ``environmental cost'' multiplied by the number of ``forest'' cells that are drilled; ``desert'' cells have zero cost.

During the game, the platform records the time required to complete the tutorial and each game, as well as a timestamped record of all user interactions, including the recommendations that are generated and the cells that are clicked.

\spara{Decision support system.}
To assist users when making the decision on where to click, we provide a Machine Learning (ML) based Decision Support System.
The DSS is trained by performing a number of random ``test drills'' to try to reconstruct the hidden oil profile.
The DSS uses an ML model based on a Lasso model fit with Least Angle Regression (LARS)~\cite{LARS_2004}; it corresponds to a Linear Model trained with an L1 prior as regularizer.
Two dimensions of the DSS that we control during 
the experiment are \textit{accuracy} and \textit{bias}.

\emph{Accuracy.}
We use a setting where three versions of the DSS are generated with respectively high, medium, and low accuracy.
The high accuracy DSS is based on an ML model trained with 20 randomly selected points, and recommends a randomly chosen cell among those predicted to have a revenue in the top 20\%.
The medium and low accuracy DSS simply generate a high accuracy recommendation and add to it circular two-dimensional Gaussian noise with $\mu=0$ and either $\sigma^2 = 3$ (small variance) or $\sigma^2 = 20$ (large variance).
The units for $\sigma$ are cells; remember each side of the map measures 32 cells.
%
%In two dimensions approximately 40\% of the recommendations will lie in a circle of radius equal to $\sigma$.
%
The medium accuracy DSS adds the small variance noise with probability 80\% and the large variance noise with probability 20\%.
The low accuracy DSS adds the small variance noise with probability 20\% and the large variance noise with probability 80\%.
These parameters are set experimentally through preliminary tests to induce a situation in which the performance of the model is not immediately obvious. 
Nevertheless, participants react to the accuracy of the DSS, as we describe in \S\ref{sec:results}.

\emph{Bias.} 
We experiment with two versions of the DSS, one providing biased predictions, and another one providing unbiased ones.
To create the biased predictions in a manner that was visible by participants, we train the biased DSS to optimize only for income, i.e., ignoring the ``environmental cost.''
We communicate bias to users, when present, by stating that the recommendation takes into account only the oil yield, but disregards the costs.
In the unbiased scenario, the DSS is trained to optimize income minus cost.
In the low-bias scenario, the DSS ignores (does not consider it during the learning phase) an environmental cost of 20\% of the maximum oil yield.
In the high-bias scenario, the DSS ignores a cost that is 40\% of the maximum oil yield.

A sequence of recommendations is pre-computed so that users get similar recommendations during their games.
The only difference they might experience is that recommendations do not suggest cells that have already been clicked.

\subsection{Experiment variables}
\label{subsec:variables}

In this section, we describe the set of \textit{independent} and \textit{dependent} variables that we identified for our experiment. The first type of variables are controlled and changed during our experimentation in order to observe their effect on the dependent variables to be measured (e.g., the user's performance). In general, any change in the independent variables may cause a change in the dependent variables.

\emph{Independent Variables.}
As independent variables we take into consideration 
%\paragraph{Independent Variables}\label{subsec:independent-variables}
%Besides demographic variables (age, gender, country of residence, education, and professional background),
%
three levels of map difficulty (easy, medium, or hard),
and conditions with and without decision support.
When decision support is present, we also consider its accuracy (low, medium, or high) and bias (absent, low, or high).

\emph{Dependent Variables.}
%\label{subsec:dependent-variables}
%
The key dependent variable is each participant's \emph{score}, which is computed by adding the income minus cost across the three maps.
Additionally, we measure the \emph{time to complete} the three maps, and we ask participants which of the three maps they perceive as the most difficult.

We measure \emph{reliance} in two ways: implicit and explicit.
Implicitly, we measure the distance between the selected cell and the provided recommendation for each play; we interpret a short distance as more reliance.
Explicitly, we use a technology acceptance survey proposed by Hoffman et al. \cite{hoffman2019metrics}.
This survey has 8 questions that are answered on a Likert scale (1-5).
The questions, which can be found in Supplementary Material~\ref{app:technology-acceptance}, address the confidence, trust, predictability, reliability, and safety of the DSS.
A score close to the maximum (40 points) indicates a high level of acceptance of the DSS, while a score close to the minimum (0 points) indicates low acceptance.

\section{Experimental Design}
\label{sec:experimental-design}

In this section we first describe the preliminary experiment used to select the three maps with different difficulty levels (\S\ref{subsec:map-selection}); then, we provide the details on the setting of our main experiments (\S\ref{subsec:experimental-design-experiments}).

\subsection{Map Selection Experiment}
\label{subsec:map-selection}

We used a crowdsourcing-based experiment to select three maps with different levels of difficulty.
In particular, for this experiment, we generated 10 candidate maps composed of a terrain profile plus an oil yield profile.
All maps were generated with equal parameters for the Perlin-noise generators~\cite{Perlin_2002}.
In particular, we used the following parameters for the generator of each of the profiles:
The terrain profile generator uses \emph{octaves}$=9$, \emph{persistence}$=0.5$ and \emph{lacunarity}$=20$,
where as the oil yield generator was parametrized with: \emph{octaves}$=1$, \emph{persistence}$=1$ and \emph{lacunarity}$=1$.
This selection of parameters yields higher surface roughness for the terrain profiles and less surface roughness for the oil profiles.\footnote{Examples of maps generated with different ranges of parameters for the Perlin-noise generator will be available in our code release with the camera-ready version of this paper.}

A total of 120 crowdsourcing workers, 15 per map, participated in this phase.
While examining the gameplay traces (score per round), we noticed in some maps a high percentage of users that started with a high score.
We called this class of participants ``luckers'' and used the proportion of these to decide which map to select; maps with a higher proportion of ``luckers'' correspond to overly simple maps.

\begin{figure}[ht] % picture
    \centering
    \includegraphics[width=0.95\columnwidth]{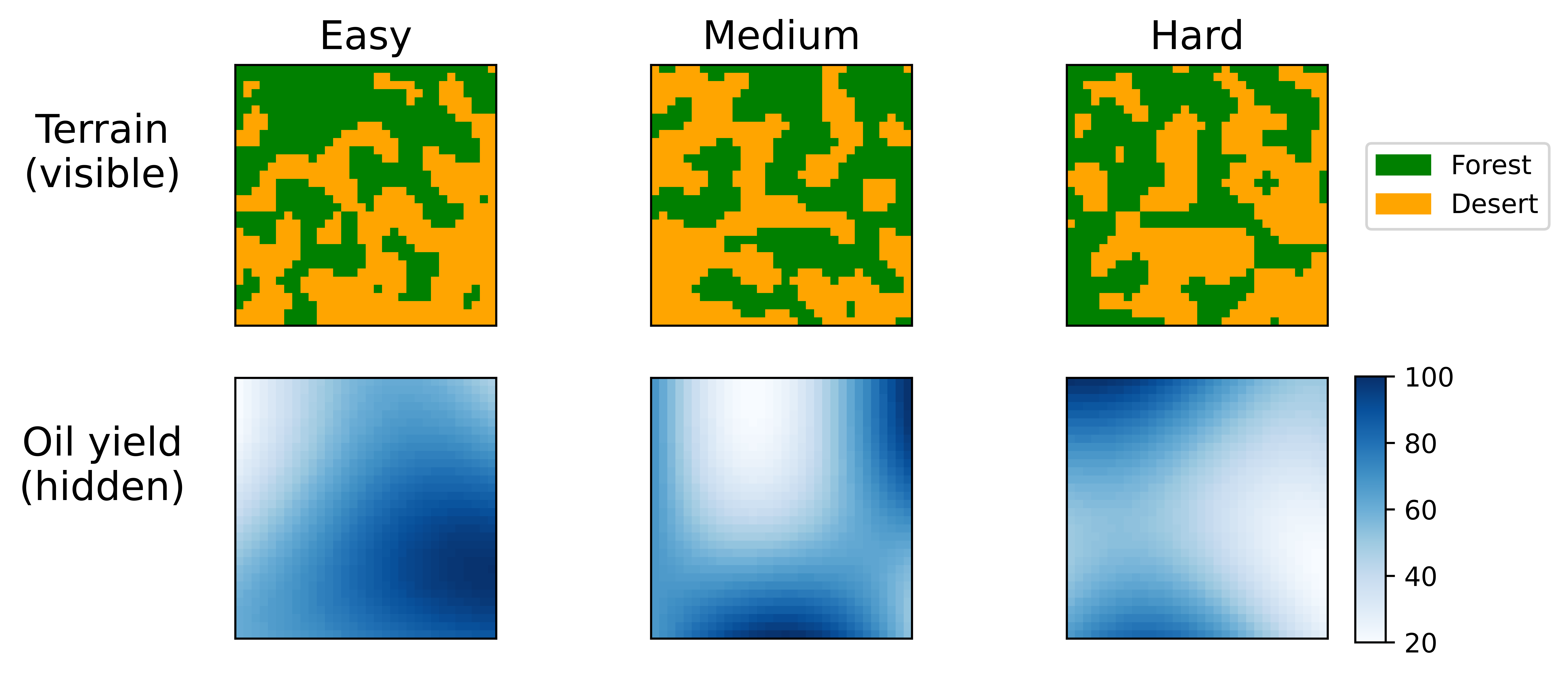}
    \caption{Selected easy, medium, and hard maps, displaying terrain (top) and oil yield distribution (bottom).
    The terrain is visible to participants; green cells represent forest, and yellow cells represent desert.
    The oil yield is hidden; darker shades indicate higher yield.
    }
    \label{fig:selected-maps}
\end{figure}

We first selected the two maps with a lower percentage of ``luckers''. The map with the lowest average score was considered  `hard'' and the other ``medium'' in terms of difficulty.
Among the rest of the maps, we selected an ``easy'' map at random among the ones in which users had the highest scores; this map has a single global optimum in the oil yield profile, and this optimum is located in a deserted area, which means users do not need to consider costs when ``drilling''.
Figure~\ref{fig:selected-maps} depicts the terrain and oil yield of the three selected maps.

The distribution of scores that we observed was aligned with map difficulty, as we show in Figure~\ref{fig:scores_distrib_by_map} (\S\ref{subsec:results-score}).
User perceptions of difficulty were also aligned with these choices, as in the exit survey when asked which one they considered to be the easiest map, the ``easy'' map was selected by 59\% of participants, the ``medium'' by 23\% of them, and the ``hard'' by the remaining 18\%.

\subsection{Main Experiments}
\label{subsec:experimental-design-experiments}

\spara{Control group.}
The control group received no machine assistance.
Users played the three maps in random order.
In total, we gathered data from 27 control group participants, testing at least three times each of the six possible map orderings.

\spara{Treatment groups.}
All the treatment groups received help from a DSS.
Here, we consider conditions combining a level of accuracy of the DSS (high, medium, or low) as described in \S\ref{subsec:platform},
with the presence or absence of bias and the amount of potential bias, as described in Table~\ref{tab:experiment3_names}.
This produced 36 (6 possible map orderings $\times$ 6 possible orderings of the DSS by levels of accuracy) possible experimental units; each one was completed by at least three different participants.
In total, 435 participants played 1,305 games.
Unique participants were recruited for each experiment separately.

\begin{table}[]
\centering\begin{tabular}{lcc}
\toprule
\multicolumn{1}{l}{\textbf{Environmental cost}} &
  \multicolumn{1}{l}{\textbf{Biased DSS}} &
  \multicolumn{1}{l}{\textbf{Unbiased DSS}} \\ \midrule
20 (low cost) &
  LB &
  LU \\ 
40 (high cost) &
  HB &
  HU \\ \bottomrule
\end{tabular}
\caption{Experimental conditions, with ``L'' representing low cost of drilling a forest cell, and ``H'' representing high cost.
LB and HB correspond to cost-unaware decision support, which yields biased suggestions.
LU and HU correspond to cost-aware decision support.}
\label{tab:experiment3_names}
\end{table}

\section{Results}
\label{sec:results}

In this section, we analyze under different experimental conditions the obtained score (\S\ref{subsec:results-score}), the time to complete the task (\S\ref{subsec:results-time}), and the reliance of participants on the DSS (\S\ref{subsec:results-reliance}).

\subsection{Score}
\label{subsec:results-score}

In this section, we observe how the score obtained by participants changes under various conditions.
%
%As a general observation, we observe that the decision support system increases the obtained score.

\spara{Map Difficulty and Decision Support.}
We compare the score obtained by participants with machine assistance with respect to the control group, to understand whether the presence of a DSS improves performance or not.

\begin{figure}[ht] % picture
   \centering
    \includegraphics[width=0.95\columnwidth]{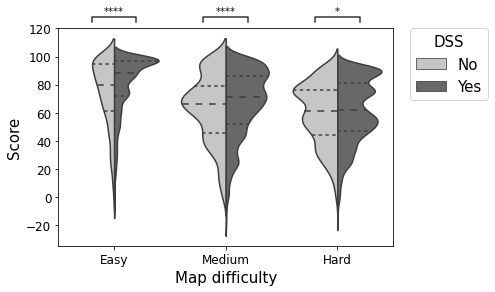}
    \caption{Distribution of scores in the three maps (easy, medium, hard), without machine assistance (left) and with machine assistance (right).
    In statistical significance tests, ``ns'' stands for no significance, and asterisks significance at: 
    % Make sure the following fits a single line:
    * ($p < 0.05$),
    ** ($p < 0.01$),
    *** ($p < 0.001$),
    **** ($p < 0.0001$).}
    %\mynote[from=Manuel]{I think in this plot the difference between groups is not easy to appreciate. Except for the Easy map, the others seems similar.}
    %\mynote[from=ChaTo]{I enlarged it a bit and perhaps it could be enlarged further by changing "Machine Assistance" to "Machine$\backslash\backslash$assistance" or "DSS" in the legend.}
    
    \label{fig:scores_distrib_by_map}
\end{figure}

Figure~\ref{fig:scores_distrib_by_map} shows the score distribution in each of the three maps, with and without machine assistance.
In this and the following figures, scores are presented per play (click), and the maximum score is 100, which is the maximum oil yield. 
Median scores per play obtained using the DSS in the easy, medium, and hard maps are respectively 88, 71, and 62 points.
Without using the DSS, these scores are respectively 80, 66, and 61 points.
Using a t-test we observe that the increase in score due to machine assistance is statistically significant at $p<0.0001$ for the easy and medium map, and at $p<0.05$ for the hard map.

\spara{Decision Support Quality and Bias.}
We evaluate the performance obtained with each level of accuracy and condition of bias of the DSS, to understand the impact that these model characteristics might have on the obtained score.

\begin{figure}[ht] % picture
    \centering
    \includegraphics[width=0.95\columnwidth]{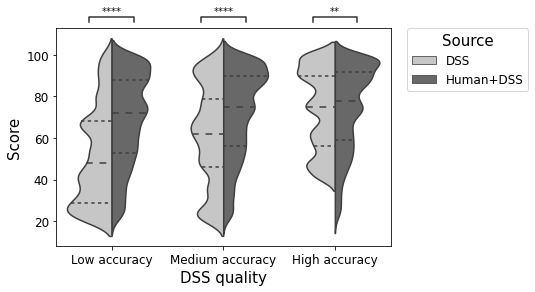}
    \caption{Score distribution by DSS quality. We compare the score obtained by human participants with machine assistance and the score that the machine would obtain. }
    \label{fig:with_DSS_scores_by_model}
    
\end{figure}

Figure~\ref{fig:with_DSS_scores_by_model} compares the distribution of scores that the DSS systems would obtain on their own (remember the DSS recommendations are randomized), against the scores obtained by participants with the help of the DSS.
From the figure, it is evident that the accuracy of the decision support system impacts the score, with more accurate systems inducing a better score.
Median scores per play obtained by participants across all three maps using the DSS are 70, 75, and 77 points using the low, medium, and high accuracy DSS respectively.
These scores are higher than what this DSS would obtain on its own: 33, 48, and 74 points respectively.
%
% 3 points (high accuracy), 13 points (medium accuracy), and 22 points (low accuracy).
Differences are statistically significant at $p<0.0001$ for the low and medium accuracy case, and at $p<0.001$ for the high accuracy case.

Furthermore, participants on their own outperform the DSS in every map, as we mentioned they obtain 80, 66, and 61 points respectively for the easy, medium, and hard maps;
in contrast, the median scores obtained by even the high-accuracy DSS are 69, 65, and 60 points respectively (figure omitted for brevity).
The median scores obtained by the medium-accuracy DSS and low-accuracy DSS are even lower.

%% EFFECT OF BIAS
Next, we consider differences in performance due to \emph{bias}.
We introduced a bias that is visible to participants by providing a biased DSS that considers only the reward but not the cost.
In contrast, the unbiased DSS considers the costs.
We considered two conditions of high terrain cost and low terrain cost (see Table~\ref{tab:experiment3_names}).

\begin{figure}[ht] % picture
    \centering
  
  \includegraphics[width=0.95\columnwidth]{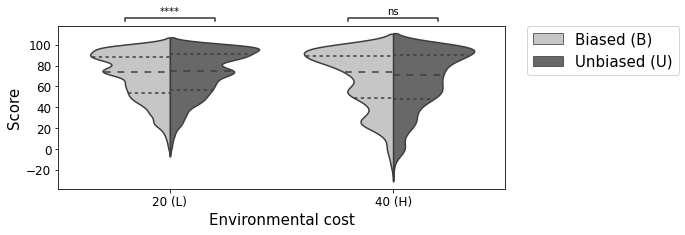}
    \caption{Score distribution using a \underline{b}iased DSS or an \underline{u}nbiased one, for the \underline{l}ow (LB vs LU) and \underline{h}igh cost (HB vs HU) conditions.}
    
    \label{fig:score_comparison_Bias_noBias}
  
\end{figure}

Figure~\ref{fig:score_comparison_Bias_noBias} shows the score distributions under the four studied conditions.
These results suggest there are some variations in score distributions, but there is no consistent increase or decrease in the median score.
Under the low-cost condition, the unbiased DSS leads to an average score per play of 75 points, while the biased DSS leads to a score of 74 points.
Under the high-cost condition, the unbiased DSS leads to a median score of 74 points, while the biased DSS to 71 points.
The difference is statistically significant in the low cost condition ($p < 0.0001$), but not in the high cost condition ($p > 0.05$).

In Supplementary Material~\ref{app:score-bias} we perform an analysis per map and per cost condition, observing that the unbiased DSS leads to higher scores in almost all cases, with the exception of the medium-difficulty map in the high-condition cost.

\spara{Probability of Bad Plays.}
We now examine the extent to which the DSS might prevent users from clicking on low-scoring cells, to understand whether the DSS,under various experimental conditions, helps the user avoid these clicks.

\begin{figure}[ht] % picture
  \centering
  \includegraphics[width=0.98\columnwidth]{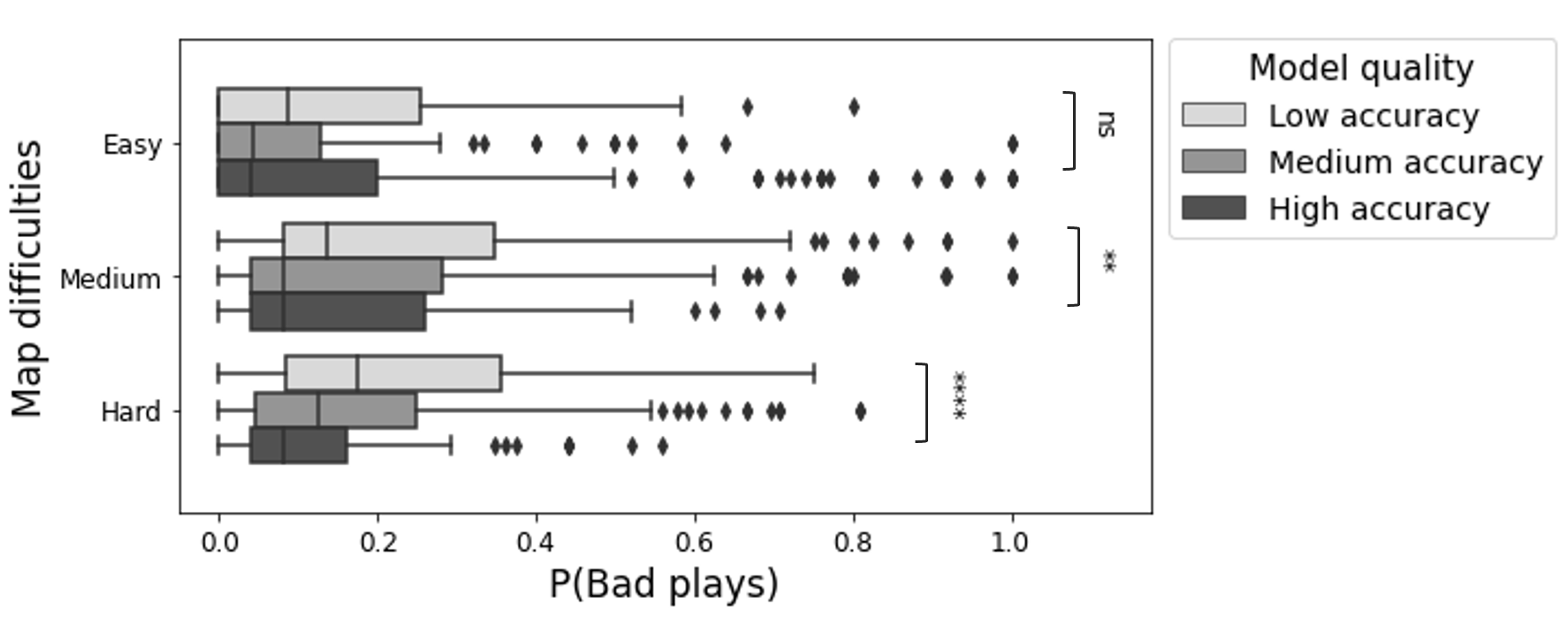}
    \caption{Probability of ``bad plays'' (below median score) under different DSS accuracy.}
  \label{fig:bad_plays_dss_quality}
\end{figure}

The distributions shown in Figure~\ref{fig:scores_distrib_by_map} suggest that the increase in performance due to the DSS is, at least in part, due to a decrease in the probability of obtaining a low score.
To study this hypothesis, we define a ``bad play'' as a click on a cell with a score lower than the median in a map.
Figure~\ref{fig:bad_plays_dss_quality} shows that the DSS reduces the probability of bad plays, and that the reduction is in general larger when the DSS is more accurate, particularly in the medium and hard maps.

\begin{figure}[ht] % picture
    \centering
    \includegraphics[width=0.8\columnwidth]{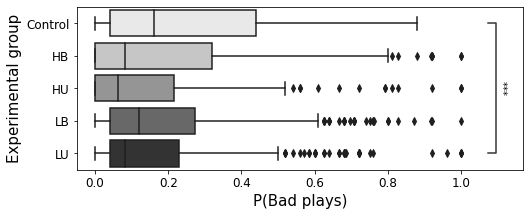}
    \caption{Probability of ``bad plays'' (below median score) under different conditions of DSS bias. HB and HU are \underline{b}iased and \underline{u}nbiased DSS, respectively, in the \underline{h}igh cost condition. LB and LU correspond to the \underline{l}ow cost condition.}
    \label{fig:bad_plays_dss_bias}
\end{figure}

Figure~\ref{fig:bad_plays_dss_bias} show that unbiased DSS (LU and HU conditions) reduce more the chances of a bad play than the biased DSS (LB and HB).

\spara{Map Ordering and Learning Effects.}
We study how the score per click obtained by participants changes as they play the game, and we observe it increases in general, which we interpret as a learning effect.
We do this first across maps/games and then within a map/game.
%\textit{Using the following analysis, we aim to understand the effect of learning. For that, we first evaluate the learning curves in each game. After, we evaluate the learning across games, assessing whether the position in which a map is presented between the three games has an effect on the obtained score}.

The experiments are designed so that players play three games in a row, with the three maps shown in random order.
Players, in general, improve their performance as they gain experience.
In all maps, the scores participants obtain when the map is in the third position are higher than the scores they obtain when the map is in the first position (figure omitted for brevity).
On average, in maps played in the first position participants obtain an average median score of 69 points whereas, in maps played in the third position, participants obtain an average median score of 79 points.
%
%For each map, we compare the scores obtained per play when it is shown in the first game with respect to the case when it is presented in the third game.
%We observe an increment of about 10 points per play on average when maps are played in the third position. 
%
We use a Kolmogorov-Smirnov (KS) two-sample test to compare them, finding that obtained differences between score distributions are statistically significant ($p<0.0001$) in each map. % 1e-14

%
%Easy
%-3.439383930319039
%KstestResult(statistic=0.14199150382134582, pvalue=1.887379141862766e-15)

%Medium
%-8.680837093639411
%KstestResult(statistic=0.17730193255866886, pvalue=4.234072813158482e-37)

%Hard
%-15.511308817215166
%KstestResult(statistic=0.3354896488436508, pvalue=4.3298697960381105e-15)

We can also compute a \emph{learning curve} for each user by concatenating the scores they obtain in each play and in each map, i.e., games are characterized as a time series composed of the timestamps and scores obtained for each of the 25 plays from each game.
To understand these learning curves, we follow a clustering-based approach that has been shown to be useful for examining learning behavior \cite{peach2019datadriven}.
This approach uses Dynamic Time Warping (DTW) to compute distances and construct a similarity matrix where values express the level of similarity between each pair of games.
Using a Relaxed Minimum Spanning Tree (RMST), we prune the weakest similarities and then use Markov Stability~\cite{delvenne2009stability,Lambiotte_2014} to obtain clusters of games with similar temporal behaviors.
Between the produced multi-scale clustering, we use the one that yields a lower number of clusters.

\begin{figure}[ht] % picture
    \centering
    \includegraphics[width=0.85\columnwidth]{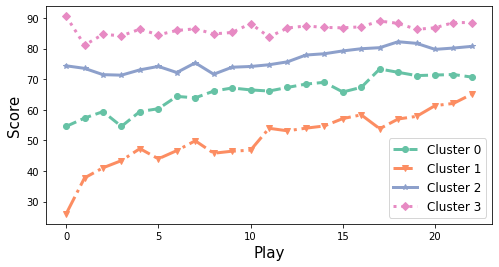}
    \caption{Centroids of clusters illustrating how the average score per play increases as players learn during each game.}
    %\mynote[from=ChaTo,to=David]{Shading of confidence intervals (or +/- one stdev) around the lines could be useful? Legend: ``Community'' $\rightarrow$ ``Cluster''; X-axis label: ``Time'' $\rightarrow$ ``Play''}
    \label{fig:communities_with_DSS}
\end{figure}

Figure~\ref{fig:communities_with_DSS} describes the centroids of the obtained clustering, composed of 4 clusters.
This figure suggests that all users tend to improve their score as their progress, and that this improvement is to a large extent dependent on their initial scores.

\spara{Exploration and exploitation behavior.}
Finally, we consider the extent to which exploration/exploitation behavior may be affected by experimental conditions.

\begin{figure}[h]
\includegraphics[width=.6\columnwidth]{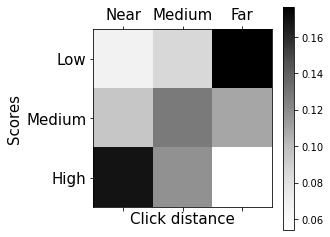}
 \caption{Exploration/exploitation behavior and performance.
 We associate clicks near each other (first column) with exploitation, and clicks far from each other (third column) with exploration.
 %
 %Left: without DSS. Right: with DSS.
 %
 Values indicate percentage of plays.}
 \label{fig:exploration_explotation}
 \end{figure}

This game requires participants to balance \emph{exploration}, i.e., seeking new high-yield areas, and \emph{exploitation}, i.e., reaping the rewards from high-yield areas already found.
This behavior is, to some extent, observable.
Two consecutive clicks near each other can be interpreted as exploitation, while consecutive clicks far from each other can be interpreted as exploration.
We are particularly interested in the extent to which these happen in the presence of a DSS, and on whether they are fruitful in the sense of leading to high-score plays.
Figure~\ref{fig:exploration_explotation} compares the euclidean distance between two consecutive clicks (in the X-axis) and the obtained score (in the Y-axis).
For clarity, we group distances and scores.
We observe that exploration (third column, ``far'') often leads to low scores, while exploitation (first column, ``near'') often leads to high scores.

Examining the score profile of several games, we observe that many participants spend the initial plays locating high-yield areas, and then switch to exploiting those.
%
%to some extent the DSS already provides some visibility about certain areas, which are assumed by the player to be high-yield given they are recommended.
%
Comparing the proportion of exploration/exploitation behaviour described in Figure~\ref{fig:exploration_explotation} with that obtained without the DSS (figure omitted for brevity), we do not find significant differences.
Both matrices differ by less than 0.02 in terms of RMSE (Root-Mean-Square Error). % MSE=0.00036, hyphenating as in the Wikipedia page
Furthermore, we do not observe statistically significant differences in the performance of a given type of play (near/medium/far click).
Binomial tests focused on determining whether the probability of obtaining a high score changes for a given click distance, between the no-DSS and the DSS conditions, yield p-values larger than 0.05.
This suggests that the DSS does not lead to a change in strategy by making participants more willing to explore or more willing to exploit, but instead makes both exploration and exploitation more efficient, proportionally.

\subsection{Time}
\label{subsec:results-time}
In this section, we evaluate the completion time across games.

\begin{figure}[ht] % picture
    \centering
    \includegraphics[width=0.95\columnwidth]{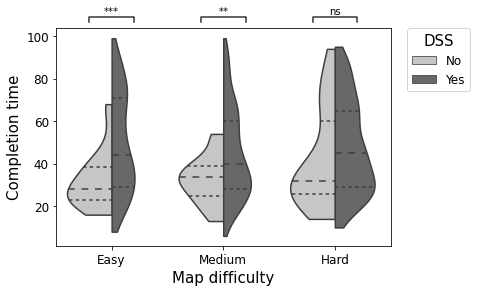}
    \caption{Completion time distribution by map. For representation purposes, outlier games taking more than 100 seconds have been removed.}
    \label{fig:time_distrib_by_map}
\end{figure}

For games played without a DSS, we measured a completion time of $37\pm18$ seconds (average and standard deviation) per map.
Given there are 25 rounds, this means users click on a cell roughly once every 1.5 seconds.
In games with the DSS, the completion time was $55 \pm 40$ seconds per map.
This corresponds to about one cell clicked every 2.2 seconds.
It is clear that the DSS induces a longer completion time, about 50\% longer.
The distribution of completion times is shown in Figure~\ref{fig:time_distrib_by_map}, where we also observe that completion time is in general correlated with map difficulty, and harder maps take longer to complete.

\subsection{Reliance on Machine Assistance}
\label{subsec:results-reliance}

In this section, we use to approaches to measure the extent to which participants are willing to rely on the DSS.

\begin{figure}[ht] % picture
    \centering
    \includegraphics[width=0.49\textwidth]{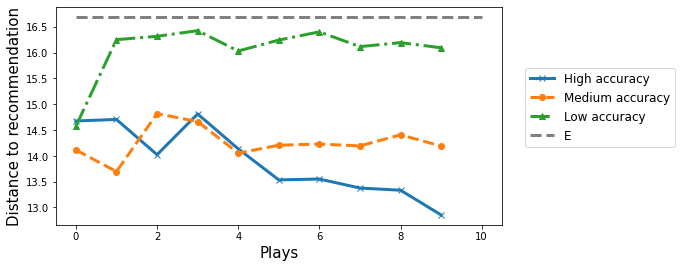}
    \caption{Average distance between recommendation and selected cell by model quality. The horizontal line (E) is the expected distance between two random points in a 32x32 grid}
    %\mynote[from=ChaTo]{Perhaps add a horizontal line at 0.52140 x 32 = 16.6848 , and indicate in the caption ``The horizontal line is the expected distance between two random points in a 32x32 grid.''?}
    \label{fig:users_reliance}
\end{figure}

In the \emph{implicit} approach, we measure the distance between the recommended point and the cell selected by the user.
This is depicted in Figure~\ref{fig:users_reliance}.
Users seem to correctly account for the accuracy of the system and rely less on the low accuracy DSS than on the medium or high accuracy DSS.
Indeed, in the low accuracy condition, after the initial play users basically ignore the DSS -- note that the expected distance between two randomly-chosen points in a $32 \times 32$ square\footnote{Analytically, in a unit square this is $\left(2 + \sqrt{2} + 5 \ln\left(\sqrt{2} + 1\right)\right)/15 \approx 0.52140\dots$} is approximately  $0.52 \cdot 32 = 16.6$, which is close to what we observe in this condition.
% Source for the 0.52 figure is:
% https://mindyourdecisions.com/blog/2016/07/03/distance-between-two-random-points-in-a-square-sunday-puzzle/
%
In contrast, in the medium accuracy condition and particularly in the high accuracy condition, they tend to click closer to the recommended point.

%\todo[from=ChaTo, to=David]{I would like to see Figure~\ref{fig:users_reliance} for LB vs LU and HB vs HU, please.}
\begin{comment}
\begin{figure}[ht] % picture
    \centering
    \includegraphics[width=0.45\textwidth]{Figures/reliance_in_recommendations_LB-LC.png}
    \caption{Average distance between recommendation and selected cell by model quality. The horizontal line (E) is the expected distance between two random points in a 32x32 grid. LB and LU experiments}
    %\mynote[from=ChaTo]{Perhaps add a horizontal line at 0.52140 x 32 = 16.6848 , and indicate in the caption ``The horizontal line is the expected distance between two random points in a 32x32 grid.''?}
    \label{fig:users_reliance_LB_LC}
\end{figure}

\begin{figure}[ht] % picture
    \centering
    \includegraphics[width=0.45\textwidth]{Figures/reliance_in_recommendations_HB-HC.png}
    \caption{Average distance between recommendation and selected cell by model quality. The horizontal line (E) is the expected distance between two random points in a 32x32 grid. HB and HU experiments}
    %\mynote[from=ChaTo]{Perhaps add a horizontal line at 0.52140 x 32 = 16.6848 , and indicate in the caption ``The horizontal line is the expected distance between two random points in a 32x32 grid.''?}
    \label{fig:users_reliance_HB_HC}
\end{figure}
\end{comment}

We also ask users to tell us \emph{explicitly} their acceptance of the DSS by using the Technology Acceptance Survey described in \S\ref{subsec:variables}, which yields a score between 0 (complete rejection) and 40 (complete acceptance).
\begin{figure}[ht] % picture
    \centering
    \includegraphics[width=0.35\textwidth]{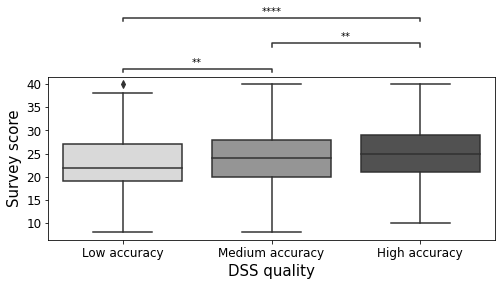}
    \caption{Results of the exit survey on technology acceptance, by DSS accuracy.}
    \label{fig:TAM_survey}
    %\mynote[from=ChaTo]{Could it be clearer to say ``DSS quality'' instead of ``Model quality''?}
\end{figure}
Results, shown in Figure~\ref{fig:TAM_survey}, indicate that most users have an intermediate level of acceptance (20-30 points out of 40) and median acceptance differs by less than 5  points across accuracy conditions.
However, high acceptance is more likely in the high and medium accuracy conditions than in the low accuracy condition, and differences in the distribution of acceptance are statistically significant at $p<0.01$.

\section{Discussion}
\label{sec:discussion}

Our research question was ``\emph{How do the characteristics of a decision support system impact human performance, time to completion, and reliance?}''

First, we observe that the platform and the experiments we designed allow experimentation on the stated variables, as participants respond in an observable manner to aspects such as the accuracy and bias of the received support.
We also observe that completing a game does not take much time, and that all participants seem to learn how to increase their score after a few interactions.
The decision support we provide, when operating on its own, has lower performance than an average human participant; however, the combination of human and machine intelligence in this game can outperform both the machine and the human acting on their own.
%
%The set-up we have designed and implemented requires no prior knowledge or experience, and seems to provide a good platform for experimentation.
% , and hence this game can be used for experimentation with large populations.

Second, our experimental results suggest that, in this context, to a large extent participants behave rationally with respect to the accuracy of the DSS.
Despite us intentionally adding noise to the DSS recommendations, participants respond to the average accuracy of the DSS,
following more closely the high accuracy DSS than the medium accuracy DSS, and to a large extent ignoring the low accuracy DSS.
This result is aligned with previous work in which participants correctly calibrated their reliance according to DSS accuracy (e.g., \cite{Yin2019,Yu_2016}).

%This result is aligned with findings from previous work. For example, \cite{Yin2019} shows that a lower observed accuracy shows less agreement, while a higher observed accuracy increases the agreement between the human and the system, regardless of the level of stated accuracy.  Additionally, in \cite{Yu_2016}, authors compared trust as a function of different levels of a DSS accuracy, showing that experiment participants correctly calibrated their trust according to machine accuracy.

Third, participants' response to the presence of bias in this DSS is found to be small and inconsistent, as they obtain slightly higher scores with the unbiased DSS in the presence of low cost but slightly lower scores with the unbiased DSS when the cost is higher.
It might be possible that a larger bias might lead to a more directly observable effect, but we remark that the high cost, biased condition introduces a bias of 40 points per play (out of 100), which is fairly substantial.

Fourth, we observe that the presence of a DSS leads users to take about 50\% longer to complete each task.
Previous work in decision-making for other tasks also observed an increased completion time (e.g., \cite{portela2022comparative, Panayiotis_2021, langer2021changing}).

We observed no statistically significant differences among participants of different socio-demographics (gender, age, country, education, professional background) in terms of any of the dependent variables studied (score, time to completion, or reliance).

A result that is somewhat unexpected is that participants, despite apparently ignoring the low accuracy DSS, in many cases express a moderate acceptance of it in the exit survey.
Previous work has found that a wrong recommendation is less penalized if the final task performance is not harmed~\cite{Yang2016}, but in our case participants indeed obtain a lower score with the low-accuracy DSS.
This suggests that, in practice, a malfunctioning DSS might not be detectable by users.\footnote{This was humorously dubbed the ``Functional Indeterminacy Theorem'' by John Gall \cite{gall1986systemantics}.} 
This can have severe consequences when deploying a DSS without taking into account that its users might not be able to evaluate correctly the quality of the recommendations or predictions.

\section{Limitations and Future Work}
\label{sec:conclusions}

Decision-making processes in professional usage and/or in high-stakes scenarios are different from those of inconsequential decisions, such as the game we have designed.
First, feedback is rarely available immediately, and indeed the process of acquiring expertise in some domains, such as criminal justice, involves to some extent observing the consequences of decisions made years ago~\cite{holsinger2018rejoinder}.
Second, while we compensate economically participants for executing their task, this is to encourage attentiveness, and not to simulate a high-stakes situation.
Third, different professional contexts in which DSS are used (such as healthcare, human resources, criminal justice) may encourage different practices with respect to the DSS;
they may also involve people with different backgrounds, including varying degrees of numeracy and different levels of previous experience with algorithmic support.
Considering this, we believe DSS studies leading to domain-specific designs require domain-specific experimentation.
What we provide, in contrast, is a platform to explore quantitatively user response to key aspects of a DSS at a scale.

A further limitation is that while we release a platform in which parameters can be varied by researchers, we do not provide mechanisms to, for instance, predict the difficulty of a game.
We provide three test maps, and notice that varying some parameters of the map generator and of the DSS probably requires experimentally fine-tuning other parameters.
We recommend doing this empirically -- as we have done in this paper -- with a group of participants.

\spara{Future work.}
We offer an expressive platform that is useful for various types of research; indeed, we encourage researchers to use this platform and we make freely available its source code.
As future work, we would like to consider situations that induce over-reliance or under-reliance in the DSS.
We would also like to study whether communicating the accuracy and confidence of the DSS, or using other mechanisms for transparency or explainability can prevent these situations, or lead to increased user performance.
Another line of research we would like to explore is the response of participants to DSS failures, such as a sudden drop in accuracy, both in terms of how they perform and how they perceive the system.
%
% I had this idea below, inspired by previous work, but I think as it is, it might not be ethical -- ChaTo
%We would like to explore a purchasing decision scenario, in which participants can ``invest'' a portion of their future reward in ``purchasing'' a DSS, with the intention of obtaining an even higher reward due to increased accuracy.

\spara{Ethics and data protection.}
Our research was approved by our university's Ethics Review Board, including a data protection assessment.

\spara{Reproducibility.}
Our platform will be available as free software with the camera-ready version of this paper.

\bibliographystyle{acm}  
\bibliography{references}  

\begin{thebibliography}{10}

\bibitem{Acharya2018}
{\sc Acharya, A., Howes, A., Baber, C., and Marshall, T.}
\newblock {Automation reliability and decision strategy: A sequential decision
  making model for automation interaction}.
\newblock {\em Proceedings of the Human Factors and Ergonomics Society 1\/}
  (2018), 144--148.

\bibitem{Balfe2018UnderstandingIK}
{\sc Balfe, N., Sharples, S., and Wilson, J.~R.}
\newblock Understanding is key: An analysis of factors pertaining to trust in a
  real-world automation system.
\newblock {\em Human Factors 60\/} (2018), 477 -- 495.

\bibitem{Bansak2019}
{\sc Bansak, K.}
\newblock {Can nonexperts really emulate statistical learning methods? A
  comment on "the accuracy, fairness, and limits of predicting recidivism"}.
\newblock {\em Political Analysis\/} (2019), 370--380.

\bibitem{Bansal2019}
{\sc Bansal, G., Nushi, B., Kamar, E., Lasecki, W.~S., Weld, D.~S., and
  Horvitz, E.}
\newblock {Beyond Accuracy: The Role of Mental Models in Human-AI Team
  Performance}.
\newblock {\em Proceedings of the AAAI Conference on Human Computation and
  Crowdsourcing 7}, 1 (2019), 19.

\bibitem{Birhane2021}
{\sc Birhane, A.}
\newblock {Algorithmic injustice: a relational ethics approach}.
\newblock {\em Patterns 2}, 2 (feb 2021), 100205.

\bibitem{Bogen2018HelpWA}
{\sc Bogen, M., and Rieke, A.}
\newblock Help wanted: an examination of hiring algorithms, equity, and bias,
  2018.

\bibitem{Bucinca2020}
{\sc Bu{\c{c}}inca, Z., Lin, P., Gajos, K.~Z., and Glassman, E.~L.}
\newblock {Proxy tasks and subjective measures can be misleading in evaluating
  explainable AI systems}.
\newblock In {\em International Conference on Intelligent User Interfaces,
  Proceedings IUI\/} (2020), vol.~20, pp.~454--464.

\bibitem{Cummings04automationbias}
{\sc Cummings, M.~L.}
\newblock Automation bias in intelligent time critical decision support
  systems.
\newblock In {\em AIAA 3rd Intelligent Systems Conference\/} (2004), AIAA,
  pp.~2004--6313.

\bibitem{delvenne2009stability}
{\sc Delvenne, J.~C., Yaliraki, S.~N., and Barahona, M.}
\newblock Stability of graph communities across time scales, 2009.

\bibitem{Dietvorst_2014}
{\sc Dietvorst, B., Simmons, J., and Massey, C.}
\newblock Algorithm aversion: People erroneously avoid algorithms after seeing
  them err.
\newblock {\em Journal of experimental psychology. General 144\/} (11 2014).

\bibitem{Dressel2018}
{\sc Dressel, J., and Farid, H.}
\newblock {The accuracy, fairness, and limits of predicting recidivism}.
\newblock {\em Science Advances 4}, 1 (2018), 1--6.

\bibitem{LARS_2004}
{\sc Efron, B., Hastie, T., Johnstone, I., and Tibshirani, R.}
\newblock Least angle regression.
\newblock {\em The Annals of Statistics 32}, 2 (Apr 2004).

\bibitem{gall1986systemantics}
{\sc Gall, J.}
\newblock {\em Systemantics: the underground text of systems lore: how systems
  really work and especially how they fail}.
\newblock General Systemantics Press, 1986.

\bibitem{Green2020a}
{\sc Green, B.}
\newblock {The false promise of risk assessments: Epistemic reform and the
  limits of fairness}.
\newblock {\em FAT* 2020 - Proceedings of the 2020 Conference on Fairness,
  Accountability, and Transparency\/} (2020), 594--606.

\bibitem{Green2019a}
{\sc Green, B., and Chen, Y.}
\newblock {Disparate Interactions}.
\newblock In {\em Proceedings of the Conference on Fairness, Accountability,
  and Transparency\/} (New York, NY, USA, jan 2019), ACM, pp.~90--99.

\bibitem{Green2019}
{\sc Green, B., and Chen, Y.}
\newblock {The principles and limits of algorithm-in-the-loop decision making}.
\newblock {\em Proceedings of the ACM on Human-Computer Interaction 3}, CSCW
  (2019).

\bibitem{Green2020}
{\sc Green, B., and Chen, Y.}
\newblock {Algorithmic risk assessments can alter human decision-making
  processes in high-stakes government contexts}, 2020.

\bibitem{Grgic-Hlaca2019}
{\sc Grgic-Hlaca, N., Engel, C., and Gummadi, K.}
\newblock Human decision making with machine assistance: An experiment on
  bailing and jailing, 10 2019.

\bibitem{Hoff2015}
{\sc Hoff, K.~A., and Bashir, M.}
\newblock {Trust in automation: Integrating empirical evidence on factors that
  influence trust}.
\newblock {\em Human Factors 57}, 3 (2015), 407--434.

\bibitem{hoffman2019metrics}
{\sc Hoffman, R.~R., Mueller, S.~T., Klein, G., and Litman, J.}
\newblock Metrics for explainable ai: Challenges and prospects, 2019.

\bibitem{holsinger2018rejoinder}
{\sc Holsinger, A.~M., Lowenkamp, C.~T., Latessa, E., Serin, R., Cohen, T.~H.,
  Robinson, C.~R., Flores, A.~W., and VanBenschoten, S.~W.}
\newblock A rejoinder to dressel and farid: New study finds computer algorithm
  is more accurate than humans at predicting arrest and as good as a group of
  20 lay experts.
\newblock {\em Fed. Probation 82\/} (2018), 50.

\bibitem{jahanbakhsh2020experimental}
{\sc Jahanbakhsh, F., Cranshaw, J., Counts, S., Lasecki, W.~S., and Inkpen, K.}
\newblock An experimental study of bias in platform worker ratings: The role of
  performance quality and gender.
\newblock In {\em Proceedings of the 2020 CHI Conference on Human Factors in
  Computing Systems\/} (2020), pp.~1--13.

\bibitem{Angwin_2016}
{\sc Julia~Angwin, Jeff~Larson, S.~M., and Lauren~Kirchner, P.}
\newblock Machine bias: There’s software used across the country to predict
  future criminals. and it’s biased against blacks., 05 2016.

\bibitem{Lai2019a}
{\sc Lai, V., and Tan, C.}
\newblock {On human predictions with explanations and predictions of machine
  learning models: A case study on deception detection}.
\newblock {\em FAT* 2019 - Proceedings of the 2019 Conference on Fairness,
  Accountability, and Transparency\/} (2019), 29--38.

\bibitem{Lambiotte_2014}
{\sc Lambiotte, R., Delvenne, J.-C., and Barahona, M.}
\newblock Random walks, markov processes and the multiscale modular
  organization of complex networks.
\newblock {\em IEEE Transactions on Network Science and Engineering 1}, 2
  (2014), 76--90.

\bibitem{langer2021changing}
{\sc Langer, M., K{\"o}nig, C.~J., and Busch, V.}
\newblock Changing the means of managerial work: effects of automated decision
  support systems on personnel selection tasks.
\newblock {\em Journal of business and psychology 36}, 5 (2021), 751--769.

\bibitem{larus2018computers}
{\sc Larus, J., Hankin, C., Carson, S.~G., Christen, M., Crafa, S., Grau, O.,
  Kirchner, C., Knowles, B., McGettrick, A., Tamburri, D.~A., et~al.}
\newblock When computers decide: European recommendations on machine-learned
  automated decision making, 2018.

\bibitem{Lee2004}
{\sc Lee, J.~D., and See, K.~A.}
\newblock {Trust in Automation: Designing for Appropriate Reliance}.
\newblock {\em Human Factors: The Journal of the Human Factors and Ergonomics
  Society 46}, 1 (jan 2004), 50--80.

\bibitem{Lin2020}
{\sc Lin, Z.~J., Jung, J., Goel, S., and Skeem, J.}
\newblock {The limits of human predictions of recidivism}.
\newblock {\em Science Advances 6}, 7 (feb 2020), 1--8.

\bibitem{lotlikar_2021}
{\sc Lotlikar V.~S., S. N. . G.~A.}
\newblock Brain tumor detection using machine learning and deep learning: A
  review.
\newblock {\em Current medical imaging\/} (2021).

\bibitem{Mallari2020}
{\sc Mallari, K., Inkpen, K., Johns, P., Tan, S., Ramesh, D., and Kamar, E.}
\newblock {Do I Look Like a Criminal? Examining how Race Presentation Impacts
  Human Judgement of Recidivism}.
\newblock In {\em Proceedings of the 2020 CHI Conference on Human Factors in
  Computing Systems\/} (New York, NY, USA, apr 2020), ACM, pp.~1--13.

\bibitem{Mason764}
{\sc Mason, W., and Watts, D.~J.}
\newblock Collaborative learning in networks.
\newblock {\em Proceedings of the National Academy of Sciences 109}, 3 (2012),
  764--769.

\bibitem{Mukerjee_2002}
{\sc Mukerjee, A., Biswas, R., Kalyanmoy, Y., Amrit, D., and Mathur, P.}
\newblock Multi-objective evolutionary algorithms for the risk-return trade-off
  in bank loan management.
\newblock {\em International Transactions in Operational Research 9\/} (03
  2002).

\bibitem{peach2019datadriven}
{\sc Peach, R.~L., Yaliraki, S.~N., Lefevre, D., and Barahona, M.}
\newblock Data-driven unsupervised clustering of online learner behaviour,
  2019.

\bibitem{Perlin_2002}
{\sc Perlin, K.}
\newblock Improving noise.
\newblock In {\em Proceedings of the 29th Annual Conference on Computer
  Graphics and Interactive Techniques\/} (New York, NY, USA, 2002), SIGGRAPH
  ’02, Association for Computing Machinery, p.~681–682.

\bibitem{portela2022comparative}
{\sc Portela, M., Castillo, C., Tolan, S., Karimi-Haghighi, M., and Pueyo,
  A.~A.}
\newblock A comparative user study of human predictions in algorithm-supported
  recidivism risk assessment, 2022.

\bibitem{Panayiotis_2021}
{\sc Smeros, P., Castillo, C., and Aberer, K.}
\newblock Sciclops.
\newblock {\em Proceedings of the 30th ACM International Conference on
  Information \& Knowledge Management\/} (Oct 2021).

\bibitem{stevenson2021algorithmic}
{\sc Stevenson, M.~T., and Doleac, J.~L.}
\newblock Algorithmic risk assessment in the hands of humans, 2021.

\bibitem{Suresh2020}
{\sc Suresh, H., Lao, N., and Liccardi, I.}
\newblock {Misplaced Trust: Measuring the Interference of Machine Learning in
  Human Decision-Making}, 2020.

\bibitem{Tatasciore2019}
{\sc Tatasciore, M., Bowden, V.~K., Visser, T. A.~W., and Loft, S.}
\newblock Should we just let the machines do it? the benefit and cost of action
  recommendation and action implementation automation.
\newblock {\em Human Factors\/} (2019).

\bibitem{Tolan2018}
{\sc Tolan, S.}
\newblock Jrc digital economy working paper 2018-10 fair and unbiased
  algorithmic decision making : Current state and future challenges, 2018.

\bibitem{Yang2016}
{\sc Yang, X.~J., Wickens, C.~D., and H{\"{o}}ltt{\"{a}}-Otto, K.}
\newblock {How users adjust trust in automation: Contrast effect and hindsight
  bias}.
\newblock {\em Proceedings of the Human Factors and Ergonomics Society\/}
  (2016), 196--200.

\bibitem{Yin2019}
{\sc Yin, M., Vaughan, J.~W., and Wallach, H.}
\newblock {Understanding the effect of accuracy on trust in machine learning
  models}.
\newblock {\em Conference on Human Factors in Computing Systems -
  Proceedings\/} (2019), 1--12.

\bibitem{yin_2019}
{\sc Yin, M., Wortman~Vaughan, J., and Wallach, H.}
\newblock Understanding the effect of accuracy on trust in machine learning
  models.
\newblock In {\em Proceedings of the 2019 CHI Conference on Human Factors in
  Computing Systems\/} (New York, NY, USA, 2019), CHI ’19, Association for
  Computing Machinery, p.~1–12.

\bibitem{Yu_2016}
{\sc Yu, K., Berkovsky, S., Conway, D., Taib, R., Zhou, J., and Chen, F.}
\newblock Trust and reliance based on system accuracy.
\newblock In {\em Proceedings of the 2016 Conference on User Modeling
  Adaptation and Personalization\/} (New York, NY, USA, 2016), UMAP '16,
  Association for Computing Machinery, p.~223–227.

\bibitem{Zhang2020}
{\sc Zhang, Y., Liao, Q.~V., Bellamy, R. K.~E., {Vera Liao}, Q., and Bellamy,
  R. K.~E.}
\newblock {Efect of confidence and explanation on accuracy and trust
  calibration in AI-assisted decision making}.
\newblock {\em FAT* 2020 - Proceedings of the 2020 Conference on Fairness,
  Accountability, and Transparency\/} (jan 2020), 295--305.

\end{thebibliography}

\appendix

\section*{Supplementary Material}

\section{Technology Acceptance Survey Questions}
\label{app:technology-acceptance}

The technology acceptance survey~\cite{hoffman2019metrics} contains the following questions:
\begin{enumerate}
    \item \textit{I am confident  in the algorithm (DSS). I feel it works well.}
    \item \textit{The outputs of the algorithm are very predictable.}
    \item \textit{The tool is very  reliable.  I can count on it to be correct all the time.}
    \item \textit{I feel safe that when I rely on the algorithm I will get the right answers.}
    \item \textit{The algorithm is efficient in that it works very quickly.}
    \item \textit{I am wary (suspicious/distrustful) of the algorithm.}
    \item \textit{The app can perform the task better than a novice human user.}
    \item \textit{I like using the system for decision making.}
\end{enumerate}

\section{Score distribution by bias condition and enviroment cost}
\label{app:score-bias}

\begin{figure}[ht!]
\includegraphics[width=.6\columnwidth]{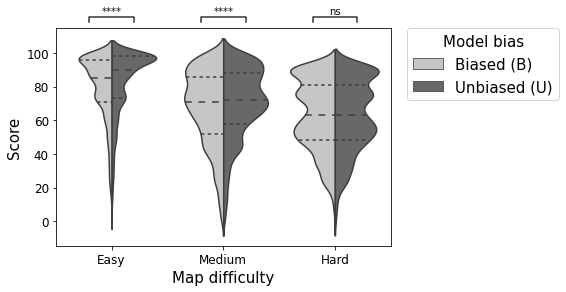}
 \caption{Score comparison of \underline{b}iased (LB) and \underline{u}nbiased (LU) DSS under a \underline{l}ow cost condition.}
 \label{fig:score_distribution_LB_LC}
 \end{figure}

\begin{figure}[ht!]
\includegraphics[width=.6\columnwidth]{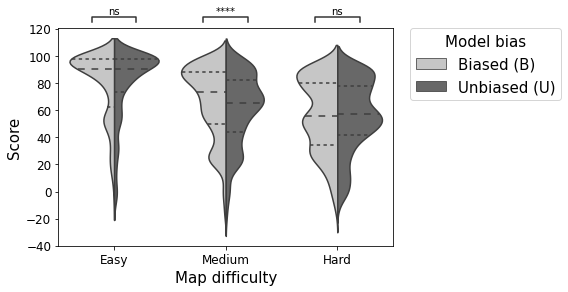}
\caption{Score comparison of \underline{b}iased (HB) and \underline{u}nbiased (HU) DSS under a \underline{h}igh cost condition.} \label{fig:score_distribution_HB_HC}
 \end{figure}

These results complement the ones in \S\ref{subsec:results-score} by comparing biased and unbiased DSS per map and under different cost conditions: high cost (HB vs HC, Figure~\ref{fig:score_distribution_HB_HC}) and low cost (LB vs LC, Figure~\ref{fig:score_distribution_LB_LC}).

%%%%%%%%%%%%%%%%
% REMOVE BEFORE SUBMITTING
%%%%%%%%%%%%%%%%
%\input{Sections/99.Removed}

\end{document}